\pdfoutput=1

\documentclass[11pt]{article}

\usepackage[]{ACL2023}

\usepackage{times}
\usepackage{latexsym}

\usepackage[T1]{fontenc}

\usepackage[utf8]{inputenc}

\usepackage{microtype}

\usepackage{inconsolata}

\usepackage{microtype}
\usepackage{color}
\usepackage{bm}
\usepackage{amssymb}
\usepackage{amsfonts}
\usepackage{amsmath}
\usepackage{footmisc}
\usepackage{multirow}
\usepackage{multicol}
\usepackage{caption}
\usepackage{subcaption}
\usepackage{latexsym}
\usepackage{mflogo}
\usepackage{amsthm}
\usepackage{graphicx} 
\usepackage{booktabs}
\usepackage{xspace}
\usepackage{adjustbox}
\usepackage{newfloat}
\usepackage{listings}
\usepackage[ruled,linesnumbered]{algorithm2e}

\newcommand{\paratitle}[1]{\vspace{1.5ex}\noindent\textbf{#1}}
\newcommand{\ie}{\emph{i.e.,}\xspace}

\newcommand{\eg}{\emph{e.g.,}\xspace}

\newcommand{\ignore}[1]{}
\newcommand{\methodname}{\textsc{VAWI}\xspace}
\definecolor{greentext}{rgb}{0.2,0.7,0}

\title{Visually-augmented Pretrained Language Models for NLP Tasks \\without Images}

\author{
    \textbf{Hangyu Guo\textsuperscript{{1}}\thanks{\llap{}\:\:\:Equal contributions.},
	        Kun Zhou\textsuperscript{{3},{4}}\footnotemark[1],
	        Wayne Xin Zhao\textsuperscript{{2},{4}}\thanks{\llap{}\:\:\:Corresponding authors. },
	        Qinyu Zhang\textsuperscript{{1}}~\and
                Ji-Rong Wen\textsuperscript{{2},{3},{4}}}\\
         \textsuperscript{1}School of Electronics and Information Engineering, Harbin Institute of Technology~(Shenzhen).\\
	\textsuperscript{2}Gaoling School of Artificial Intelligence, Renmin University of China.\\
	\textsuperscript{3}School of Information, Renmin University of China.\\
	\textsuperscript{4}Beijing Key Laboratory of Big Data Management and Analysis Methods.\\
        \texttt{hyguo0220@gmail.com, francis\_kun\_zhou@163.com} \\
	\texttt{batmanfly@gmail.com, zqy@hit.edu.cn, jrwen@ruc.edu.cn}
}

\begin{document}
\maketitle
\begin{abstract}

Although pre-trained language models~(PLMs) have shown impressive performance by text-only self-supervised training, they are found lack of visual semantics or commonsense.
Existing solutions often rely on explicit images for visual knowledge augmentation (requiring time-consuming retrieval or generation), and they also conduct the augmentation for the whole input text, without considering whether it is actually needed in specific inputs or tasks.   
To address these issues, we propose a novel \textbf{V}isually-\textbf{A}ugmented fine-tuning approach that can be generally applied to various PLMs or NLP tasks, \textbf{W}ithout using any retrieved or generated \textbf{I}mages, namely \textbf{VAWI}.
Experimental results show that our approach can consistently improve the performance of BERT, RoBERTa, BART, and T5 at different scales, and outperform several competitive baselines on ten tasks.
Our codes and data are publicly available at~\url{https://github.com/RUCAIBox/VAWI}.

\end{abstract}

\section{Introduction}

Recent years have witnessed the success of pre-trained language models (PLMs)~\citep{qiu2020pre,LLMSurvey}, such as GPT-3~\citep{brown2020language} and T5~\citep{t5}, in a variety of natural language process~(NLP) tasks. Since these PLMs are mostly trained on text-only corpus via self-supervised pre-training, they have been shown lack of visual commonsense~\citep{liu2022things} and real-world knowledge~\citep{zhang2022visual}. 
As a result, PLMs can't well solve visually related language tasks~\footnote{In this work, we mainly focus on text-only NLP tasks that may benefit from external visual information, rather than visual-language tasks involving images.}, \eg answering the color and size of common things, especially those requiring complex commonsense knowledge. 

To alleviate this problem, existing works mainly enhance PLMs by infusing visual information.
Typically, given a text input, these studies firstly augment the visual information from retrieved or generated images about the input and then leverage their visual representations to improve PLMs on NLP tasks. Such an approach leads to \emph{visually-augmented pre-trained language models (VaLMs)}, where they adopt either visually-augmented pre-training~\citep{tan2020vokenization,wang2022visually} or visually-augmented fine-tuning~\citep{lu2022imagination}.
Despite the effectiveness, there are two major shortcomings in these methods.
First, these methods often rely on pre-learned complementary retrievers or generators, and also require time-consuming inference to retrieve or generate proper images that are paired with the input.
The above costly conditions largely limit the applicability of these approaches. 
Second, the retrieved or generated images are inevitable to involve irrelevant or redundant visual information. If simply integrating them, the original text representations might be affected.
Increasing evidence shows that the visual information is not always useful for NLP tasks~\citep{dai2022enabling}, and sometimes leads to performance degradation.

Considering these issues, we aim to develop a more efficient and effective way to visually augment the PLMs and the solution is  twofold: 

~~$\bullet$ Firstly, we don't explicitly produce (retrieve or generate) the images but instead generate visually-aligned representations of the text on-the-fly. 
Recent studies~\citep{radford2021learning,jia2021scaling} have shown that the vision-language pre-trained models~(VL-PTMs) can well learn the alignment between the representations of texts and images from large-scale text-image pairs. Thus, our idea is to employ the output representations of a text from VL-PTMs' text encoders as a surrogate for the visual representations of related images.

Such a way is simple and efficient: we can only keep the text encoder of a VL-PTM to produce the visually-aligned representations of texts, 
getting rid of the complicated image retrieval or generation process.
It is widely recognized that there is a large semantic gap between different modalities~\citep{liang2022mind}. 
Our method can alleviate this issue to some extent since the visual augmentations are derived from the text representation itself.

~~$\bullet$ Secondly, instead of directly feeding visual augmentations into the PLM, we propose to use the augmented visual information only when it is actually required. In fact, for a text input of a NLP task,  PLMs are not always hungry for the visual background knowledge to effectively understand it, especially for visually-irrelevant expressions. 
Unlike previous works which inject visual information into a text~\citep{tan2020vokenization,wang2022visually} from the whole, we consider identifying 
\emph{visually-hungry words} (those that require visual knowledge to derive complete semantics) from the text input, and only infuse the visual augmentations through these trigger words. We conduct visual augmentations at the word level, because it is more flexible and controllable, considering the augmented information is often irrelevant or noisy. 

To this end, in this paper, we propose a general \textbf{V}isually-\textbf{A}ugmented fine-tuning approach to improving PLMs for NLP tasks \textbf{W}ithout \textbf{I}mages, namely \textbf{VAWI}.
Our approach consists of three ingredients, namely visually-hungry words extraction, visual knowledge augmentation, and visually-enhanced fine-tuning.
Given the text input from a NLP task, we first extract the visually-hungry words~(VH-words) from the input sentence.
As the annotations of VH-words are generally unavailable, we propose three strategies to automatically extract the VH-words, relying on the syntax trees, attention distributions of VL-PTMs, and an adaptive learnable module, respectively.
Then, based on the extracted VH-words, we leverage the text encoder of CLIP~\citep{radford2021learning} (being fixed in our approach), a VL-PTM that has been pre-trained on millions of text-image pairs, to encode the VH-words for obtaining their visually-aligned representations.
Finally, we infuse the visually-aligned representations into PLMs, and consider the general and parameter-efficient fine-tuning strategies for small and large PLMs, respectively.

To verify the effectiveness of our framework \textbf{VAWI}, we test it on four PLMs~(\ie BERT, BART, RoBERTa, and T5) at different scales~(\ie 110M, 340M, 3B), and conduct extensive experiments in natural language understanding, commonsense reasoning, and text generation tasks.
Experimental results show that our \textbf{VAWI} can boost the performance of these PLMs significantly, \ie 3.11\%, 2.54\%, and 2.16\% absolute improvements on the commonsenseQA task using RoBERTa-base, RoBERTa-large, and T5-3b, respectively.
Besides, \textbf{VAWI} can outperform (or be on par with) several competitive baselines that adopt complicated visually-augmented methods.
\section{Related Work}

\paratitle{Pre-trained Language Models.}
Recent years have witnessed the success of pre-trained language models~(PLMs)~\citep{Devlin2019BERTPO, gpt}.
After pre-trained on the large-scale corpus, PLMs can be fine-tuned on multiple NLP tasks and achieve remarkable performance.
However, since PLMs are just pre-trained with text-only data, they may suffer from the reporting bias problem~\citep{gordon2013reporting, paik2021world, zhang2022visual}, where the frequency distribution of visual commonsense in the text may not fully reflect the real-world distribution of the commonsense. 
Existing works have also found that such a problem can not be well addressed by enlarging the model or pre-training corpus~\citep{paik2021world, zhang2022visual}.
In this work, we aim to alleviate this problem by adding visual knowledge on PLMs during fine-tuning.

\paratitle{Vision-Language Pre-Trained Models.}
To better accomplish the vision-language tasks, vision-language pre-trained models~(VL-PTMs)~\citep{su2019vl,lu2019vilbert} become a hot point in recent years, which require large-scale image-text pairs for pre-training. 
Existing VL-PTMs fall into two categories based on the way of modeling vision-language interaction. 
The first category of models~\citep{lu2019vilbert,li2021align} adopts an explicit vision-language interaction layer to fuse the text embeddings and image features.
These models are more suitable to capture fine-grained semantic interactions between vision and language.
The second category of models~\citep{radford2021learning,jia2021scaling} incorporates separate encoders to model the vision and language information, and relies on pre-training tasks (\eg cross-modal contrastive learning) to align their representations into the same latent space.
Such a way is capable of producing enriched single-modal representations.

\paratitle{Visually-Augmented Language Model.}
To introduce visual information into PLMs, visually-augmented language model~(VaLM)~\citep{wang2022visually} has become an emerging research topic.
Existing VaLMs can be categorized into visually-augmented pre-training and fine-tuning.
Visually-augmented pre-training approaches~\citep{tan2020vokenization,iNLG} continually pre-train PLMs with the retrieved visual information related to input tokens or sentences and also revise the masked language model task for better capturing the visual semantics.
Visually-augmented fine-tuning method~\citep{lu2022imagination} introduces the visual information into PLMs during fine-tuning.
These methods also leverage the image retrieval or generation models to augment the visual information and design a special fusion module to inject it into PLMs.
However, existing VaLM approaches mostly need to retrieve or generate visual information for utilization.
Such a way is time-consuming, and may involve unrelated or noisy information into PLMs, leading to performance degradation.
In this work, we aim to first detect the visually-hungry words from the text, and then utilize a VL-PTM to generate their visually-aligned representations without the usage of external images or generation models.
As a comparison, our approach is more flexible and efficient to leverage visual information for enhancing text-based PLMs.

\section{Method} \label{method}
    \begin{figure*}[t]
    \centering
    \includegraphics[width=\textwidth]{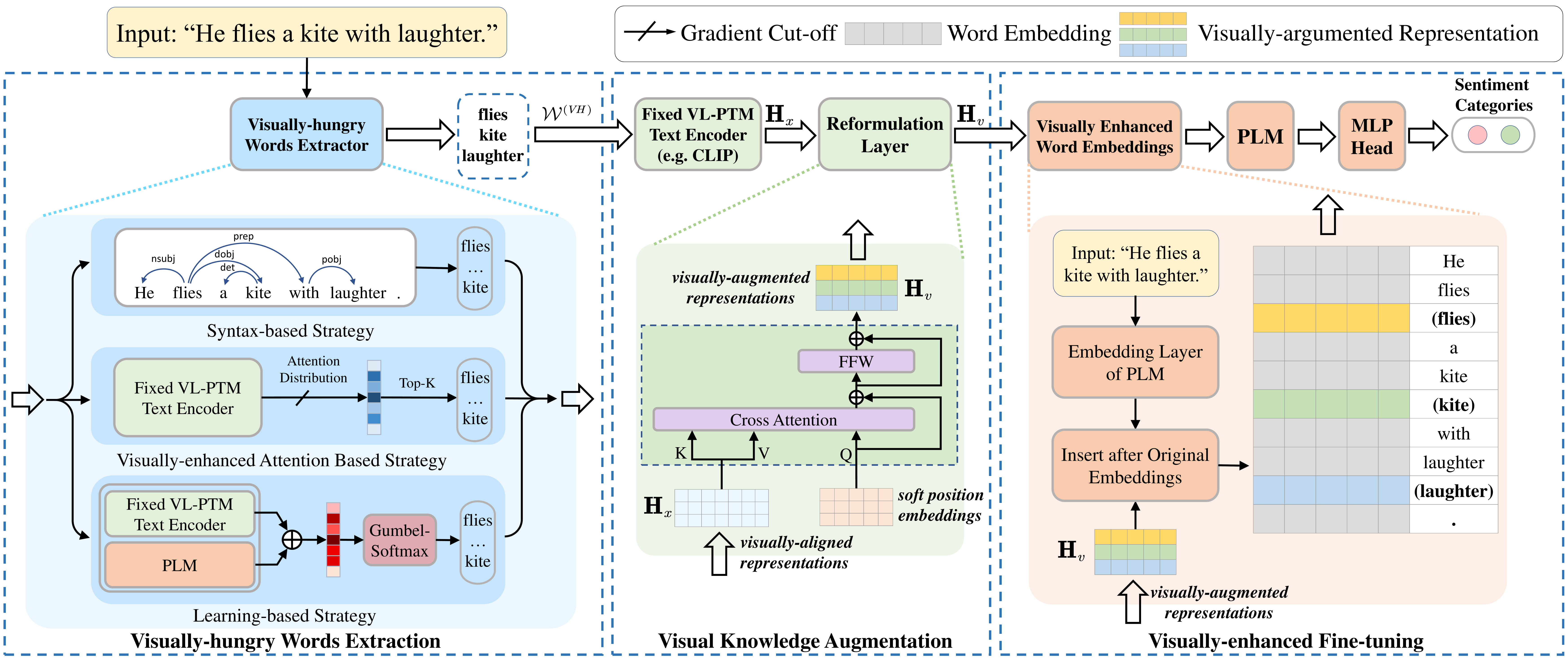}
    \caption{The illustration of our \methodname approach, consisting of visually-hungry words extraction, visual knowledge augmentation and visually-enhanced fine-tuning.}
    \label{fig:model}
\end{figure*}

In this section, we firstly introduce the task setting, and then describe our proposed visual augmentation approach for infusing visual knowledge into PLMs during fine-tuning.

\subsection{Task Setting and Solution Overview}
This work aims to improve the fine-tuning performance of pre-trained language models~(PLMs) on NLP tasks by leveraging the related visual information without images.
For a NLP task, a set of $n$ labeled texts $\{\langle x_{i}, y_{i} \rangle\}$ are available, where $x_{i}$ is the $i$-th text data consisting of a sequence of words, denoted as $x_{i}=\{w_{1}, w_{2}, ..., w_{m}\}$, and $y_{i}$ is the ground-truth output, which can be a discrete label~(classification), a continuous value~(regression) or a text sequence~(generation).
    
To solve the target task, we assume that a text-based PLM is given (either for understanding or generation). 
Let $f$ denote a PLM parameterized by $\theta_{\text{PLM}}$ that has already been pre-trained on general-purpose large-scale text data. 
Given the labeled training data, we can train the PLM using a specific loss function (\eg cross-entropy loss) and further solve the target task.   
However, existing works~\citep{tan2020vokenization,zhang2022visual} have revealed that PLMs may be unaware of visual knowledge that is not explicitly mentioned in the pre-trained text-only data (\eg the shape of coins and the color of the sky), leading to the lack of world commonsense and generating wrong statements.

In this work, we focus on devising an efficient and effective way to infuse such visual knowledge into PLMs during fine-tuning.
Our approach is based on \emph{visually-hungry words} (abbreviated as \emph{VH-words}), which require visual information to derive complete semantic representations.
The overall illustration of our approach is shown in Figure~\ref{fig:model}.
Given the input text $x_i$ and its label $y_i$, we first detect and extract a set of VH-words. 
Then, we adopt a visual knowledge augmentation module to enhance the visual background knowledge of their tokens and generate their visually-aligned representations. 
Finally, we infuse the visually-aligned text representations into the PLM to improve its fine-tuning performance, where we consider both the general fine-tuning of small PLMs and the parameter-efficient fine-tuning of large-scale PLMs.

\subsection{Visually-Hungry Words Extraction}
In our approach, visually-hungry words~(VH-words) are the trigger units for visual augmentations, requiring visual knowledge for deriving complete semantic representations (\eg color, shape, and object).
Therefore, we propose to first detect the VH-words from the input text, and then inject the proper visual knowledge that they are hungry for into the PLM.
However, the annotations about VH-words are generally not available in NLP datasets.
To address this problem, we devise three different strategies to extract the VH-words from the input text, including two feature-based strategies based on syntax tree and attention distribution of PLMs, and a learnable model-based strategy.

\paragraph{Syntax-based Strategy.}
In natural language, entity words and descriptive words usually convey more visual semantics than others. 
For example, for the sentence ``\emph{He is eating a \underline{green apple}}'', where underlined words are more related to visual semantics.
Such words are mostly nouns or adjectives in the input text, which can be detected by syntactic analysis.
Therefore, we design a rule-based strategy that leverages the syntactic information for VH-words extraction.
Concretely, we first delete all stop words in a text and then adopt an open-resource toolkit \textsc{spacy}~\footnote{\url{https://spacy.io/}} to convert the input text into a syntax dependency tree.
Based on the syntax tree, we extract the words that have a particular part of speech~(POS), \eg nouns or adjectives, as the VH-words denoted by $\mathcal{W}^{(VH)}$.
In this way, we can efficiently extract the VH-words from input text by using a fast parser toolkit.

\paragraph{Visually-enhanced Attention Based Strategy.}
The attention-based strategy utilizes the attention distribution of a VL-PTM to detect the VH-words.
Since VL-PTMs~\citep{radford2021learning} are pre-trained on large-scale image-text pairs, their text encoders can focus more on the words corresponding to some specific visual concepts in an image, which are likely to be VH-words.
Inspired by it, we use the attention scores calculated by the text encoder of VL-PLMs to select the VH-words.
Specifically, we adopt the text encoder of CLIP~\citep{radford2021learning}, a VL-PTM that has been pre-trained on millions of image-text pairs, to help extract the VH-words.
As CLIP adopts an autoregressive GPT-2 model as the text encoder, we calculate the average attention scores between each token and the ``\texttt{[EOS]}'' token on the self-attention layer, denoted as $s_{w_{i}}$.
Then, we select the top-$K$ ranked words according to $\{ s_{w_{i}} \}$ as the VH-words $\mathcal{W}^{(VH)}$.

\paragraph{Learning-based Strategy.} Considering that diverse PLMs and NLP tasks may be hungry for different complementary visual information, we devise a learning-based strategy that can adaptively extract VH-words according to task requirements.
Concretely, we add a parameterized VH-words extractor layer for the PLM, which can be updated by gradient-based optimization algorithms to fit the need for some specific task.
Given the input text $x_i$, we first leverage the PLM and a text encoder of a VL-PTM~(\ie CLIP~\citep{radford2021learning}) to produce the contextualized representations of the contained words in  $x_i$.
Then, we concatenate the representations of each word from the two models and utilize a MLP layer to obtain the score $s_{w_{i}}$:
\begin{equation}
    s_{w_{i}}=\text{MLP}([\mathbf{h}^{(P)}_{w_{i}};\mathbf{h}^{(V)}_{w_{i}}])
\end{equation}
where $\mathbf{h}^{(P)}_{w_{i}}$ and $\mathbf{h}^{(V)}_{w_{i}}$ are the output word representations from the PLM and VL-PTM, respectively, and scores~$s_{w_{i}}$ are calculated by the learned model based on the supervision information from downstream tasks.
Based on the scores of all words, we incorporate the gumbel-softmax function~\citep{jang2016categorical} to extract the top-$k$ words as the VH-words in a differentiable way.
In this way, the gradients of the fine-tuned tasks can be back-propagated to the extractor layer, which learns to adaptively select the more suitable VH-words.

\subsection{Visual Knowledge Augmentation}
Existing works~\citep{lu2022imagination,wang2022visually} mainly utilize image retrieval or generation module to augment related visual knowledge.
Such a way is time-consuming and may also involve noisy images.
Inspired by recent works that show the effective visual-language alignment in VL-PTMs~\citep{radford2021learning,li2021align}, we utilize the visually-aligned text encoders to generate the visual augmentation representations of VH-words.
As the text encoders have been aligned to the image encoders during pre-training, their output textual representations can be used as surrogates of visual augmentations based on real images related to the input text.
As will be shown in experiments (Section 4), this approach is not only efficient but very effective for downstream NLP tasks.

Based on the extracted VH-words, we first add a prefix text in the image caption style before the VH-words, \eg ``\emph{a photo of: }'', to compose the input text $x'$.
Then, we utilize the text encoder of CLIP~\citep{radford2021learning} to encode $x'$ and obtain the contextualized word representations as the visually-aligned representations $\mathbf{H}_x \in \mathbb{R}^{k\times d}$, where $k$ is the sequence length of $x'$ and $d$ is the embedding size.
Next, we incorporate a reformulation layer to aggregate and strengthen the visually-aligned representation~$\mathbf{H}_x$ into the visually-augmented representations of these VH-words.
As the positions of the VH-words vary from sentence to sentence, we design a position-aware attention mechanism in the reformulation layer to inject position information into $\mathbf{H}_x$ for obtaining the visual representation of each VH-word. 
Specifically, we first leverage a soft position embedding matrix $\mathbf{E}\in \mathbb{R}^{l\times d}$ to reserve the position information of VH-words, where $l$ is the number of VH-words. Then, we perform the cross-attention between it and the visual representations as:
\begin{align}
    \mathbf{Q} &= \mathbf{E},~~\mathbf{K} = \mathbf{H}_{x}\mathbf{W}^{K} + \bm{b}^{K}, \label{eq-q}\\  
    \mathbf{V} &= \mathbf{H}_{x}\mathbf{W}^{V} + \bm{b}^{V}, \label{eq-kv}\\
    \mathbf{H}_{v} &= \text{softmax}(\frac{\mathbf{Q} \mathbf{K}^\top}{\sqrt{d}})\mathbf{V}, \\
    \mathbf{H}_{v}^\top &= [\mathbf{h}_{1},\mathbf{h}_{2},...,\mathbf{h}_{l}], \label{eq-vhwords}
\end{align}
where~$\mathbf{h}_{i}\in \mathbb{R}^{d},~\mathbf{K},\mathbf{V}\in \mathbb{R}^{k\times d}$.~$\mathbf{H}_{v} \in \mathbb{R}^{l\times d}$ is the obtained visually-augmented representations of VH-words, which is leveraged for augmenting the visual knowledge of the PLM.
$\mathbf{h}_{i}$ is the visual representation of the $i$-th VH-word in $\mathcal{W}^{(VH)}$.
Note that in Eq.~\ref{eq-q} and~\ref{eq-kv}, we adopt an efficient way that only uses the position information to set the \emph{query} matrix $\mathbf{Q}$, 
and the visual semantics are mainly captured and injected through the \emph{key} and \emph{value} matrices.

\subsection{Visually-Enhanced Fine-tuning}
After obtaining the visually-augmented representations of VH-words (\ie $\mathbf{H}_{v}$ in Eq.~\ref{eq-vhwords}), we propose a visually-enhanced fine-tuning strategy to inject the captured visual knowledge. 
Here, we consider two cases: (1) full-parameter fine-tuning for small PLMs, and (2) parameter-efficient prompt-tuning for large-scale PLMs.
Before introducing the learning method, we simply review the parameters of our approach, consisting of the parameters in the underlying PLM ($\Theta_{plm}$), the VL-PTM ($\Theta_{vlp}$) and the parameters of the reformulation layer ($\Theta_{ref}$). 
Note that we will always fix $\Theta_{vlp}$ in our approach.

\paragraph{Fine-tuning for Small PLMs.}
For small PLMs, we can perform full-parameter fine-tuning, which updates both $\Theta_{plm}$ and $\Theta_{ref}$.
Specifically, given the visually-augmented representations $\mathbf{H}_{v}$ of VH-words, we directly incorporate them into the embedding layer of the PLM.
For each VH-word, we insert its visually-augmented representation after the original word embedding, to leverage the visual semantics to enrich the word representations.

\paragraph{Prompt-tuning for Large-Scale PLMs.}
For large-scale PLMs, we fix the parameters in it, \ie $\Theta_{plm}$, and employ a parameter-efficient prompt-tuning way to optimize it on downstream NLP tasks.
Concretely, given the visually-augmented representations $\mathbf{H}_{v}$ of VH-words, we directly insert them before the input representations of every layer of PLMs.
Then, following the typical prompt-tuning paradigm~\citep{li2021prefix}, we only tune the parameters of the reformulation layer (\ie $\Theta_{ref}$) as the soft prompts to adapt all the model into the fine-tuning task.

Our approach can be generally applied to various PLMs (\eg BERT~\citep{Devlin2019BERTPO}, BART~\citep{Lewis2020BARTDS}, T5~\citep{t5}) and  NLP tasks (natural language understanding and text generation).  Unlike other complicated visually-augmented methods~\citep{tan2020vokenization,wang2022visually}, it is more efficient, without the explicit need of external images or generation model; and meanwhile, it only introduces a small number of parameters (Eq.~\ref{eq-kv}), which are easier to learn.

\section{Experiments}
\label{sec:experiments}

\begin{table*}[htbp]
    \small
    \centering
    \resizebox{\textwidth}{!}{
    \begin{tabular}{c c c c c c c c c c c c}
    \toprule
    \textbf{Base Model}  & \textbf{Method} & \textbf{SST-2} & \textbf{QNLI}  & \textbf{QQP} & \textbf{MNLI} & \textbf{MRPC} & \textbf{STS-B} &\textbf{Avg.} \\
    \midrule
    CLIP                 &+None  & 73.3          & 74.5          & 72.8          & 68.4          & 74.3          & 73.8          & 72.85\\
    BLIP                 &+None  & 76.3          & 77.4          & 78.8          & 72.5          & 77.8 & 76.4          & 76.53\\
    ALBEF$_{14M}$        &+None  & 78.9 & 78.2 & 79.4 & 73.4 & 76.5          & 77.5 & 77.31\\
    \midrule
    \multirow{6}{*}{BERT$_{base}$} 
    & +None               & 89.3          & 87.9          & 87.2          & 79.4          & 81.7          &84.4               &84.98 \\
    & +VOKEN                  & 92.2          & 88.6          & 88.6          & 82.6          & 83.5          & 86.0          & 86.83\\
    & +iACE                  & 91.7          & 88.6          & 89.1          & 82.8          & 85.8          & 86.6          & 87.43\\
    & +\methodname-SBS                    & \textbf{92.9} & 88.4          & 89.6          & 82.2          & 85.5          & 86.9          & 87.58\\
    & +\methodname-VABS          & 92.7          & 88.9          & 89.5          & 82.7          & \textbf{85.8} & \textbf{87.2} & \textbf{87.80}\\
    & +\methodname-LBS             & 92.4          & \textbf{89.1} & \textbf{89.7} & \textbf{83.0} & 85.6          & 86.9          & 87.78\\
    \midrule
    \multirow{6}{*}{RoBERTa$_{base}$}
    & +None               & 89.2          & 87.5          & 86.2          & 79.0          & 81.4          & 85.4          & 84.78\\
    & +VOKEN                  & 90.5          & 89.2          & 87.8          & 81.0          & 87.0          & 86.9	         & 87.06\\
    & +iACE                   & 91.6          & 89.1          & \textbf{87.9} & 82.6          & 87.7          & 86.9          & 87.63\\
    & +\methodname-SBS                    & 91.4	      & 89.4	        & 87.7	   & 82.2	     & 88.2	       & 87.7	         & 87.76\\
    & +\methodname-VABS                    & \textbf{91.7} & 89.1	        & \textbf{87.9} & \textbf{82.6} & 88.3	       & 88.1	         & 87.95\\
    & +\methodname-LBS             & 91.6          & \textbf{90.6} & \textbf{87.9} & 82.4	     & \textbf{88.5}	& \textbf{88.3} & \textbf{88.21}\\
    \bottomrule
    \end{tabular}}
    \caption{Performance comparison of different methods on NLU tasks, the \textbf{BEST} results are highlighted in bold. 
    \emph{+None} denotes that we directly fine-tune the backbone without adding visual information.
    SBS, VABS, and LBS represent using the syntax-based strategy, visually-enhanced attention based strategy, and learning-based strategy in our approach, respectively.
    The results of VOKEN and iACE on GLUE are reported from \citet{lu2022imagination}.} 
    \label{tablemain:glue}
\end{table*}

\subsection{Experimental Setup}
\paratitle{Datesets.}
\label{dataset}
We conduct experiments on four types of tasks.
(1) Natural Language Understanding~(NLU): we extract 6 datasets from the GLUE benchmark~\citep{wang2018glue}; 
(2) Commonsense reasoning: we select CommonsenseQA~\citep{csqa}, a 5-way multiple choice QA dataset that requires commonsense knowledge;
(3) Text generation: we select CommonGen~\citep{lin2019commongen}, a constrained text generation task about generative commonsense reasoning.
(4) Cross-modal reasoning: we select SNLI-VE~\citep{xie2019visual}, to evaluate the capacity of predicting whether the image semantically entails the text.

\paratitle{Baseline Models.}
We compare our approach with the following baselines, including pre-trained language models~(PLMs), visual-language pre-trained models~(VL-PTMs), and visually-augmented pre-trained language modes~(VaLMs).
(1) \textbf{PLMs}: We choose BERT~\citep{Devlin2019BERTPO}, RoBERTa~\citep{Liu2019RoBERTaAR}, BART~\citep{Lewis2020BARTDS}, T5~\citep{t5} as the PLM backbones, and directly fine-tune them as baselines.
(2) \textbf{VL-PTMs}: We select ALBEF~\citep{li2021align}, BLIP~\citep{li2022blip}, and CLIP~\citep{radford2021learning}, which have been pre-trained on large-scale image-text pairs.
(3) \textbf{VaLMs}: we select VOKEN~\citep{tan2020vokenization} and iACE~\citep{lu2022imagination}, which introduce the visual information into PLMs by pre-training on retrieved images and fine-tuning on generated images, respectively.

\paratitle{Implementation Details.}
\label{implementation}
We implement all methods based on Huggingface Transformers~\citep{huggingface}.
For all baselines, we set their hyper-parameters according to their papers.
In our approach, we leverage the text encoder of CLIP~(ViT-B/32) to implement the learnable model-based VH-words extractor and generate the visual representations of VH-words in the visual knowledge augmentation module. 
The hidden size of visual representations is set to 512.
For different NLP tasks, we tune the number of visually hungry words in \{2, 3, 4, 5\}.
During fine-tuning, we perform parameter-efficient tuning on T5-3b and BART-Large, and full-parameter tuning on other PLMs.
For all tasks and all backbones, we utilize Adam as the optimizer, set the learning rate to 2e-5, weight decay to 0.01, and a linear warmup for the first 6\% steps.
For GLUE, GommonGen, and SNLI-VE datasets, we fine-tune our model for 3 epochs with a batch size of 32.
For CommonsenseQA, we tune our model for 10 epochs with a batch size of 32.
We use the cross-entropy loss for classification and the mean squared error loss for regression.

\subsection{Main Experimental Results}

\begin{table*}[ht]
    \centering
    \small
    \resizebox{\textwidth}{!}{
    \begin{tabular}{cccccccccc}
        \toprule
            \multirow{2}{*}{\textbf{Base Model}}& \multirow{2}{*}{\textbf{Method}}
            &\multicolumn{4}{c}{CommonsenseQA-3k} & \multicolumn{4}{c}{CommonsenseQA}\\
            \cmidrule(lr){3-6} \cmidrule(lr){7-10}
            & & 5\% & 10\% & 20\% & 100\% & 5\% & 10\% & 20\% & 100\%\\
            \midrule
            \multirow{3}{*}{RoBERTa$_{base}$} 
            &+None                 & 41.88            & 46.04            & 50.58          & 61.88          &44.88           &50.04           &57.08           &67.90        \\
            &+Images      & 42.37            & 48.09            & 52.81          & 64.22          &45.72           &51.17           &58.96           &69.64         \\
            &+\methodname-SBS       & \textbf{42.94}   & \textbf{49.27}   & \textbf{53.97} & \textbf{65.10} &\textbf{46.51}  &\textbf{52.44}  &\textbf{59.87}  &\textbf{71.01} \\
            \midrule
            \multirow{3}{*}{RoBERTa$_{large}$} 
            &+None                  & 48.39            & 56.30            & 59.06         & 74.19          &51.24           & 59.95       & 65.52            & 76.65 \\
            &+Images       & 49.55            & 57.78            & 61.29         & 75.61          &52.18           & 60.93       & 66.08            & 78.39        \\
            &+\methodname-SBS        & \textbf{50.27}   & \textbf{58.17}   & \textbf{62.22}& \textbf{76.54} &\textbf{52.98}  & \textbf{61.97} &\textbf{67.40} & \textbf{79.19} \\
            \midrule
            \multirow{3}{*}{T5-3B} 
            &+None                 & 70.16            & 73.02            & 75.04          & 81.81          &71.99           &75.27         &77.72                  &82.40\\
            &+Images      & 70.96            & 73.60            & 75.91          & 82.40          &72.87           &76.17         &78.71                  &83.64\\
            &\methodname-SBS+PET       & \textbf{71.52}   & \textbf{74.19}   & \textbf{76.49} & \textbf{83.61} &\textbf{73.58}  &\textbf{73.58}&\textbf{79.66}          &\textbf{84.56}\\
        \bottomrule
    \end{tabular}}
    \caption{Performance comparison on CommonsenseQA-3k and CommonsenseQA with different amounts of training data.
    We report the average performance on the dev set over three runs, and the \textbf{BEST} results are highlighted in bold.
    \emph{+Images} denotes that we add retrieved images about the VH-words using web search engines, and encode them via CLIP-ViT.}
    \label{tablevice1:csqa}
    \end{table*}
    
    \begin{table*}[ht]
    \small
    \centering  
    \resizebox{\textwidth}{!}{
    \begin{tabular}{cccccccc}
    \toprule
    \textbf{Method} &\textbf{Base Model}        & \textbf{BLUE-3} & \textbf{BLUE-4} & \textbf{METOR} & \textbf{Rouge-L} & \textbf{CIDER} & \textbf{SPICE} \\
        \midrule
        \multirow{4}{*}{BART-large} 
        &+None             & 42.80         & 32.42             & 31.36          & 57.57            & 16.56          & 32.94\\
        &+Images                & 42.67         & 32.67             & 32.12          & 57.46            & 16.78         & 32.81\\
        &+\methodname-SBS           & \textbf{44.56}& \textbf{34.17}    & \textbf{32.47} & \textbf{58.46}   & \textbf{17.23} & \textbf{33.67}\\
        &+\methodname-SBS+PET  & 43.12         & 33.76             & 32.20          & 58.12            & 16.91          & 33.17\\
        \midrule
        \multirow{4}{*}{T5-3b} 
        &+None                  & 45.92         & 35.92             & 33.02          & 58.57            & 17.71          & 33.51\\
        &+Images                & 45.69         & 35.50             & 33.55          & 58.94            & 17.51          & 32.91\\
        &+\methodname-SBS           & \textbf{47.67}& \textbf{37.54}    & 33.41         & \textbf{59.94}   & \textbf{18.34} & \textbf{34.67}\\
        &+\methodname-SBS+PET  & 47.40         & 37.36             & \textbf{33.71} & 59.78            & 18.18          & 34.17\\
        \bottomrule
    \end{tabular}}
    \caption{Performance comparison on CommonGen. We also show the performance of parameter-efficient tuning of our approach, denoted as \emph{+PET}.
    The \textbf{BEST} results are highlighted in bold. }
    \label{tablevice2:commongen}
    \end{table*}
    
In this part, we conduct a series of experiments on NLU, commonsense reasoning, text generation, and cross-modal commonsense reasoning tasks.

\paratitle{Evaluation on NLU Tasks.}
We present the experimental results of different methods on 6 NLU tasks in Table~\ref{tablemain:glue}.
First, we observe that VL-PTMs perform worse than PLMs, a possible reason is that they have been continually pre-trained on large-scale image-text pairs, which may cause the catastrophic forgetting problem.
Second, VaLMs (\ie VOKEN, iACE, and VAWI) achieve better performance over PLMs.
As VaLMs infuse external visual knowledge into the PLMs, they can help the PLMs better understand the background knowledge of some words (\eg color, shape, and size of objects).
Between the two VaLM baselines, iACE is slightly better.
This is because iACE is enhanced based on VOKEN and incorporates an image generation model, so it produces more visual information to utilize.
However, the generated images inevitably contain noise and redundant information, which limits the performance gain of iACE.

Finally, by comparing our approach with all baselines, it is obvious that~\methodname performs consistently better than them on the six datasets.
In our approach, we adopt an efficient and effective way that augments the visually-augmented representations using the text encoder of CLIP to encode the VH-words from the input text.
Benefiting from pre-training on large-scale image-text pairs, the text encoder of CLIP has been well aligned with the semantic space of images, so that it can generate high-quality visually-augmented representations of the VH-words to enrich them.
Such a way not only saves the costs of time and computation but also reduces the influence of inevitable noise from retrieved or generated images.
Additionally, among three VH-words extraction strategies, LBS slightly outperforms others in most NLU tasks.
The reason is that LBS incorporates a learnable model-based strategy to select the VH-words. Such a way can adaptively extract proper VH-words with the consideration of the intrinsic knowledge of the PLMs.
However, LBS will increase the computation cost due to its involved learnable VH-words extractor layer.
Therefore, for efficiency, in the following experiments, we utilize the SBS strategy in our approach for comparison.

\paratitle{Evaluation on Commonsense Reasoning Tasks.}
Following existing works~\citep{lin2019kagnet}, we also rely on a rule-based strategy to extract the examples containing visible objects, to construct a new dataset called CommonsenseQA-3K.
It consists of 2,903 and 341 examples in the training set and dev set, respectively.
Based on the CommonsenseQA and CommonsenseQA-3k, we also report the results with different amounts of training data, to further evaluate the performance of different methods in the few-shot setting.

As shown in Table~\ref{tablevice1:csqa}, we can also see that with the help of the visual information from either retrieved images or our VAWI-SBS, the performance of PLMs can be improved significantly.
It indicates that visual information is indeed helpful to improve PLMs for understanding commonsense knowledge.
Besides, our approach outperforms the method using retrieved images from search engines.
Our approach omits the image retrieval process due to its inevitably involved noise, and relies on the text encoder of CLIP to augment the visual representations.
Such a way can guarantee the relevance between the augmented visual knowledge and the text input, reducing the influence of retrieved noisy images and redundant information.
Furthermore, we also perform parameter-efficient tuning on T5-3B-encoder with our approach and boost its performance. 
It shows that our approach is able to be applied to large-scale PLMs to meet their thirst for visual information.

\begin{table}[t]
    \centering
    \small
    \begin{tabular}{ccccc}
        \toprule
            \multirow{2}{*}{\textbf{Method}} &\multicolumn{4}{c}{SNLI-VE} \\
            \cmidrule(lr){2-5} & 10\% & 20\% & 50\% & 100\% \\
            \midrule 
            ALBEF                       & 65.46 & 67.52 & 75.47 & 80.91 \\
            ALBEF+\methodname+SBS       & \textbf{65.94} & \textbf{68.23} & \textbf{76.14} & \textbf{81.64} \\
        \hline
    \end{tabular}
    \caption{Results on the test set of SNLI-VE task. The \textbf{BEST} results are highlighted in bold.}
    \label{tablevice3:snli-ve}
\end{table}

\paratitle{Evaluation on the Text Generation Task.}
As shown in previous experiments, it is useful to improve the performance of~\methodname on commonsense reasoning and nature language understanding tasks.
Here, we would like to study the effectiveness of our approach on the text generation task (\ie CommonGen) using large PLMs.
As shown in Table~\ref{tablevice2:commongen}, our model \methodname also consistently boosts the performance of BART-Large and T5-3b among all metrics.
It further shows that our approach can also improve PLMs on the text generation task.
As a comparison, we can see that the retrieved images are not very helpful and even cause performance degradation.
The reason may be that the text generation task is more sensitive to the inevitable noise from the retrieved images.
Finally, the parameter-efficient tuning strategy of our approach also achieves comparable performance with the full-parameter tuning.
It indicates that our parameter-efficient strategy is able to efficiently optimize the parameters of large-scale PLMs, and shows a promising future to apply our approach to much larger PLMs, \eg GPT-3.

\begin{table*}[t]
	\small
	\centering
        \resizebox{\textwidth}{!}{
	\begin{tabular}{ccccccc}
	\toprule
	\textbf{Source of visual representation~(Params)}   & \textbf{CSQA-3k} &\textbf{CSQA}& \textbf{SST-2}   & \textbf{QQP} &\textbf{STS-B}&\textbf{QNLI}\\
		\midrule
		Random Noise            (0M) &61.59   &66.78   &  89.13  & 86.27 & 85.13  & 87.22  \\
		RoBERTa-large       (355M)   &61.18   &67.17   &  89.43  & 86.53 & 85.60  & 87.77 \\
		T5-large-encoder    (375M)   &62.21   &67.87   &  89.71  & 86.67 & 86.40  & 87.94 \\
		T5-3b-encoder       (1500M)  &63.10   &68.42   &  90.24  & 86.96 & 86.93  & 88.21 \\
		CLIP-base           (52M)    &\textbf{65.10}   &\textbf{71.07}   & \textbf{91.41}  & \textbf{87.72} & \textbf{87.67}  & \textbf{89.40} \\
		\bottomrule
	\end{tabular}}
	\caption{Performance comparison of different sources of visual representation in our approach. The
	base model is RoBERTa-base.}
	\label{tableablation1:PLMs}
\end{table*}

\begin{table*}[t]
	\small
	\centering  
	\begin{tabular}{cccccc}
	\toprule
	\textbf{The text encoder of different VL-PTMs~(Params)}  
									 & \textbf{CSQA-3k}   & \textbf{SST-2}  & \textbf{QQP}\\
		\midrule
		Random Noise (0M)                         & 61.59              &  89.23           & 86.21  \\
		ALBEF~(110M)                        & 63.34              &  90.72           & 87.17  \\
		CLIP-base~(52M)            & 65.10              &  91.41           & 87.72  \\
		UniCL-base~(52M)            & 65.98              &  91.75           & 88.07  \\
		CLIP-large~(123M)            & \textbf{66.27}     &\textbf{92.10}    &\textbf{88.31}  \\
		\bottomrule
	\end{tabular}
	\caption{Performance comparison of visual representations from different VL-PTMs in our approach. The base model is RoBERTa-base.}
	\label{tableablation1:VL-PTMs}
\end{table*}

\paratitle{Evaluation on the Cross-modal Commonsense Reasoning Task.}
To verify the generality of our method, we further implement our~\methodname on a VL-PTM (\ie ALBEF~\cite{li2021align}), and conduct experiments on a cross-modal reasoning dataset, SNLI-VE.
Concretely we implement our approach on ALBEF by inserting the visually-augmented representations after the VH-words embeddings of the text encoder before the multimodal encoder, and keeping others unchanged.
As shown in Table~\ref{tablevice3:snli-ve}, our \methodname can also improve the performance of ALBEF using different amounts of training data.
It further shows the generality of our approach in VL-PTMs, as it can also provide rich information to enhance the text encoder of VL-PTM, helping it better perform cross-modal reasoning.

\subsection{Ablation Study}

In this part, we conduct a series of experiments to verify whether the improvement of our approach derives from the augmented visual knowledge about the VH-words. More ablation studies are shown in Appendix~\ref{appendix_ablation}.

\paratitle{The Effect of the Source of Visual Representations.} 
We first propose three variants that incorporate powerful PLMs, \ie RoBERTa-base, T5-Large, and T5-3b respectively, to replace the text encoder of CLIP in our framework.
We also replace the generated visual representations from the text encoder of CLIP with random noise, to investigate the importance of the visual representations.
As shown in Table~\ref{tableablation1:PLMs}, we can see that our approach is better than all the variants, even T5-3b with billion-scale parameters.
It indicates that CLIP-base is more effective to augment visual knowledge to improve the performance of PLMs.
Besides, our approach also outperforms the variant using random noise as the visual representation, showing the worse performance among all the variants.
It also shows the importance of visual representations, as they indeed contain the visual knowledge that the PLM is hungry for.

\paratitle{The Effect of the Stronger VL-PTMs.}  
In our work, we choose CLIP-base to enhance PLMs, as it has been pre-trained on a large-scale image-text dataset. 
Generally, a stronger VL-PTM would be more promising to further improve the performance. 
Here, we replace our CLIP-base model with some stronger VL-PTMs, \eg ALBEF~\citep{li2021align}, UniCL-base~\citep{yang2022unified}, and CLIP-large. 
Concretely, ALBEF leverages more pre-training tasks (\eg MLM, ITM, and ITC), UniCL utilizes more high-quality pre-training data, and CLIP-large increases the scale of model parameters. 
We evaluate the above variations on CSQA-3k, QQP, and SST-2, and the results are shown in Table~\ref{tableablation1:VL-PTMs}. 
We can see that UniCL and CLIP-large outperform CLIP-base. 
It indicates that the VL-PTMs with the larger scale of model parameters or more high-quality pre-training data are more capable of augmenting useful visual knowledge for PLMs. 
Considering the efficiency, CLIP-base is also a good choice in our approach, and we will investigate more proper VL-PTMs in the future.

\section{Conclusion}
\label{sec:conclusion}

In this paper, we proposed a general visually-augmented fine-tuning approach that can be applied to a variety of PLMs and NLP tasks, without using any retrieved or generated images, namely \textbf{VAWI}.
Specifically, we first identified and extracted the visually-hungry words~(VH-words) from input text via a token selector, where three different methods have been proposed, including syntax-, attention- and learning-based strategies.
Then, we adopted a fixed VL-PTM text encoder to generate the visually-augmented representations of these VH-words. 
As it has been pre-trained by visual-language alignment tasks on the large-scale corpus, it is capable of injecting visual semantics into the aligned text representations.
Finally, we transformed the visually-aligned features into visually-augmented features by reformulation layer based on VH-words, and inserted them into PLMs to enrich the visual semantics of word representations in PLMs.
Experimental results on 10 NLP tasks show that our approach can consistently improve the performance of BERT, RoBERTa, BART, and T5 at different scales, and outperform several competitive baselines significantly.
Besides, the visual prompts of our framework can also be used for parameter-efficient tuning, which can boost the performance of large language models, such as T5-3b.

\section*{Limitations}

An important limitation of our approach VAWI is the need for extracting visually-hungry words (VH-words) as the trigger to inject visual knowledge into PLMs.
In real-world applications, it is hard to obtain the annotations of VH-words. 
Therefore, we propose three VH-words extraction strategies.
However, the three strategies may be not always proper for all NLP tasks, and we rely on the experimental results to select the best one among them.
Besides, we adopt the text encoder of CLIP as the VL-PTM for generating the visually-aligned representation.
As a pre-trained model, CLIP also may contain biases learned from the pre-training corpus, which may result in improper biased prediction on some NLP tasks.

\section*{Acknowledgement}

This work was partially supported by National Natural Science Foundation of China under Grant No. 62222215, Beijing Natural Science Foundation under Grant No. 4222027, and Beijing Outstanding Young Scientist Program under Grant No. BJJWZYJH012019100020098.  Xin Zhao is the corresponding author.

\bibliography{anthology,custom}

\begin{thebibliography}{33}
\expandafter\ifx\csname natexlab\endcsname\relax\def\natexlab#1{#1}\fi

\bibitem[{Brown et~al.(2020)Brown, Mann, Ryder, Subbiah, Kaplan, Dhariwal,
  Neelakantan, Shyam, Sastry, Askell et~al.}]{brown2020language}
Tom Brown, Benjamin Mann, Nick Ryder, Melanie Subbiah, Jared~D Kaplan, Prafulla
  Dhariwal, Arvind Neelakantan, Pranav Shyam, Girish Sastry, Amanda Askell,
  et~al. 2020.
\newblock Language models are few-shot learners.
\newblock \emph{Advances in neural information processing systems},
  33:1877--1901.

\bibitem[{Dai et~al.(2022)Dai, Hou, Shang, Jiang, Liu, and
  Fung}]{dai2022enabling}
Wenliang Dai, Lu~Hou, Lifeng Shang, Xin Jiang, Qun Liu, and Pascale Fung. 2022.
\newblock Enabling multimodal generation on clip via vision-language knowledge
  distillation.
\newblock \emph{arXiv preprint arXiv:2203.06386}.

\bibitem[{Devlin et~al.(2019)Devlin, Chang, Lee, and
  Toutanova}]{Devlin2019BERTPO}
Jacob Devlin, Ming-Wei Chang, Kenton Lee, and Kristina Toutanova. 2019.
\newblock {BERT}: Pre-training of deep bidirectional transformers for language
  understanding.
\newblock In \emph{NAACL}.

\bibitem[{Gordon and Van~Durme(2013)}]{gordon2013reporting}
Jonathan Gordon and Benjamin Van~Durme. 2013.
\newblock Reporting bias and knowledge acquisition.
\newblock In \emph{Proceedings of the 2013 workshop on Automated knowledge base
  construction}, pages 25--30.

\bibitem[{Jang et~al.(2016)Jang, Gu, and Poole}]{jang2016categorical}
Eric Jang, Shixiang Gu, and Ben Poole. 2016.
\newblock Categorical reparameterization with gumbel-softmax.
\newblock \emph{arXiv preprint arXiv:1611.01144}.

\bibitem[{Jia et~al.(2021)Jia, Yang, Xia, Chen, Parekh, Pham, Le, Sung, Li, and
  Duerig}]{jia2021scaling}
Chao Jia, Yinfei Yang, Ye~Xia, Yi-Ting Chen, Zarana Parekh, Hieu Pham, Quoc Le,
  Yun-Hsuan Sung, Zhen Li, and Tom Duerig. 2021.
\newblock Scaling up visual and vision-language representation learning with
  noisy text supervision.
\newblock In \emph{International Conference on Machine Learning}, pages
  4904--4916. PMLR.

\bibitem[{Lewis et~al.(2020)Lewis, Liu, Goyal, Ghazvininejad, Mohamed, Levy,
  Stoyanov, and Zettlemoyer}]{Lewis2020BARTDS}
Mike Lewis, Yinhan Liu, Naman Goyal, Marjan Ghazvininejad, Abdelrahman Mohamed,
  Omer Levy, Veselin Stoyanov, and Luke Zettlemoyer. 2020.
\newblock Bart: Denoising sequence-to-sequence pre-training for natural
  language generation, translation, and comprehension.
\newblock In \emph{ACL}.

\bibitem[{Li et~al.(2022)Li, Li, Xiong, and Hoi}]{li2022blip}
Junnan Li, Dongxu Li, Caiming Xiong, and Steven Hoi. 2022.
\newblock Blip: Bootstrapping language-image pre-training for unified
  vision-language understanding and generation.
\newblock \emph{arXiv preprint arXiv:2201.12086}.

\bibitem[{Li et~al.(2021)Li, Selvaraju, Gotmare, Joty, Xiong, and
  Hoi}]{li2021align}
Junnan Li, Ramprasaath Selvaraju, Akhilesh Gotmare, Shafiq Joty, Caiming Xiong,
  and Steven Chu~Hong Hoi. 2021.
\newblock Align before fuse: Vision and language representation learning with
  momentum distillation.
\newblock \emph{Advances in neural information processing systems},
  34:9694--9705.

\bibitem[{Li and Liang(2021)}]{li2021prefix}
Xiang~Lisa Li and Percy Liang. 2021.
\newblock Prefix-tuning: Optimizing continuous prompts for generation.
\newblock \emph{arXiv preprint arXiv:2101.00190}.

\bibitem[{Liang et~al.(2022)Liang, Zhang, Kwon, Yeung, and Zou}]{liang2022mind}
Weixin Liang, Yuhui Zhang, Yongchan Kwon, Serena Yeung, and James Zou. 2022.
\newblock Mind the gap: Understanding the modality gap in multi-modal
  contrastive representation learning.
\newblock \emph{arXiv preprint arXiv:2203.02053}.

\bibitem[{Lin et~al.(2019{\natexlab{a}})Lin, Chen, Chen, and
  Ren}]{lin2019kagnet}
Bill~Yuchen Lin, Xinyue Chen, Jamin Chen, and Xiang Ren. 2019{\natexlab{a}}.
\newblock Kagnet: Knowledge-aware graph networks for commonsense reasoning.
\newblock \emph{arXiv preprint arXiv:1909.02151}.

\bibitem[{Lin et~al.(2019{\natexlab{b}})Lin, Zhou, Shen, Zhou, Bhagavatula,
  Choi, and Ren}]{lin2019commongen}
Bill~Yuchen Lin, Wangchunshu Zhou, Ming Shen, Pei Zhou, Chandra Bhagavatula,
  Yejin Choi, and Xiang Ren. 2019{\natexlab{b}}.
\newblock Commongen: A constrained text generation challenge for generative
  commonsense reasoning.
\newblock \emph{arXiv preprint arXiv:1911.03705}.

\bibitem[{Liu et~al.(2022)Liu, Yin, Feng, and Zhao}]{liu2022things}
Xiao Liu, Da~Yin, Yansong Feng, and Dongyan Zhao. 2022.
\newblock Things not written in text: Exploring spatial commonsense from visual
  signals.
\newblock \emph{arXiv preprint arXiv:2203.08075}.

\bibitem[{Liu et~al.(2019)Liu, Ott, Goyal, Du, Joshi, Chen, Levy, Lewis,
  Zettlemoyer, and Stoyanov}]{Liu2019RoBERTaAR}
Yinhan Liu, Myle Ott, Naman Goyal, Jingfei Du, Mandar Joshi, Danqi Chen, Omer
  Levy, Mike Lewis, Luke Zettlemoyer, and Veselin Stoyanov. 2019.
\newblock {RoBERTa}: A robustly optimized bert pretraining approach.
\newblock \emph{ArXiv}, abs/1907.11692.

\bibitem[{Lu et~al.(2019)Lu, Batra, Parikh, and Lee}]{lu2019vilbert}
Jiasen Lu, Dhruv Batra, Devi Parikh, and Stefan Lee. 2019.
\newblock Vilbert: Pretraining task-agnostic visiolinguistic representations
  for vision-and-language tasks.
\newblock \emph{Advances in neural information processing systems}, 32.

\bibitem[{Lu et~al.(2022)Lu, Zhu, Wang, Eckstein, and Wang}]{lu2022imagination}
Yujie Lu, Wanrong Zhu, Xin~Eric Wang, Miguel Eckstein, and William~Yang Wang.
  2022.
\newblock Imagination-augmented natural language understanding.
\newblock \emph{arXiv preprint arXiv:2204.08535}.

\bibitem[{Paik et~al.(2021)Paik, Aroca-Ouellette, Roncone, and
  Kann}]{paik2021world}
Cory Paik, St{\'e}phane Aroca-Ouellette, Alessandro Roncone, and Katharina
  Kann. 2021.
\newblock The world of an octopus: How reporting bias influences a language
  model's perception of color.
\newblock \emph{arXiv preprint arXiv:2110.08182}.

\bibitem[{Qiu et~al.(2020)Qiu, Sun, Xu, Shao, Dai, and Huang}]{qiu2020pre}
Xipeng Qiu, Tianxiang Sun, Yige Xu, Yunfan Shao, Ning Dai, and Xuanjing Huang.
  2020.
\newblock Pre-trained models for natural language processing: A survey.
\newblock \emph{Science China Technological Sciences}, 63(10):1872--1897.

\bibitem[{Radford et~al.(2021)Radford, Kim, Hallacy, Ramesh, Goh, Agarwal,
  Sastry, Askell, Mishkin, Clark et~al.}]{radford2021learning}
Alec Radford, Jong~Wook Kim, Chris Hallacy, Aditya Ramesh, Gabriel Goh,
  Sandhini Agarwal, Girish Sastry, Amanda Askell, Pamela Mishkin, Jack Clark,
  et~al. 2021.
\newblock Learning transferable visual models from natural language
  supervision.
\newblock In \emph{International Conference on Machine Learning}, pages
  8748--8763. PMLR.

\bibitem[{Radford et~al.(2019)Radford, Wu, Child, Luan, Amodei, and
  Sutskever}]{gpt}
Alec Radford, Jeff Wu, Rewon Child, David Luan, Dario Amodei, and Ilya
  Sutskever. 2019.
\newblock Language models are unsupervised multitask learners.

\bibitem[{Raffel et~al.(2020)Raffel, Shazeer, Roberts, Lee, Narang, Matena,
  Zhou, Li, and Liu}]{t5}
Colin Raffel, Noam~M. Shazeer, Adam Roberts, Katherine Lee, Sharan Narang,
  Michael Matena, Yanqi Zhou, Wei Li, and Peter~J. Liu. 2020.
\newblock Exploring the limits of transfer learning with a unified text-to-text
  transformer.
\newblock \emph{ArXiv}, abs/1910.10683.

\bibitem[{Su et~al.(2019)Su, Zhu, Cao, Li, Lu, Wei, and Dai}]{su2019vl}
Weijie Su, Xizhou Zhu, Yue Cao, Bin Li, Lewei Lu, Furu Wei, and Jifeng Dai.
  2019.
\newblock Vl-bert: Pre-training of generic visual-linguistic representations.
\newblock \emph{arXiv preprint arXiv:1908.08530}.

\bibitem[{Talmor et~al.(2019)Talmor, Herzig, Lourie, and Berant}]{csqa}
Alon Talmor, Jonathan Herzig, Nicholas Lourie, and Jonathan Berant. 2019.
\newblock Commonsenseqa: {A} question answering challenge targeting commonsense
  knowledge.
\newblock In \emph{Proceedings of the 2019 Conference of the North American
  Chapter of the Association for Computational Linguistics: Human Language
  Technologies, {NAACL-HLT} 2019, Minneapolis, MN, USA, June 2-7, 2019, Volume
  1 (Long and Short Papers)}, pages 4149--4158.

\bibitem[{Tan and Bansal(2020)}]{tan2020vokenization}
Hao Tan and Mohit Bansal. 2020.
\newblock Vokenization: Improving language understanding with contextualized,
  visual-grounded supervision.
\newblock \emph{arXiv preprint arXiv:2010.06775}.

\bibitem[{Wang et~al.(2018)Wang, Singh, Michael, Hill, Levy, and
  Bowman}]{wang2018glue}
Alex Wang, Amanpreet Singh, Julian Michael, Felix Hill, Omer Levy, and Samuel~R
  Bowman. 2018.
\newblock Glue: A multi-task benchmark and analysis platform for natural
  language understanding.
\newblock \emph{arXiv preprint arXiv:1804.07461}.

\bibitem[{Wang et~al.(2022)Wang, Dong, Cheng, Song, Liu, Yan, Gao, and
  Wei}]{wang2022visually}
Weizhi Wang, Li~Dong, Hao Cheng, Haoyu Song, Xiaodong Liu, Xifeng Yan, Jianfeng
  Gao, and Furu Wei. 2022.
\newblock Visually-augmented language modeling.
\newblock \emph{arXiv preprint arXiv:2205.10178}.

\bibitem[{Wolf et~al.(2020)Wolf, Debut, Sanh, Chaumond, Delangue, Moi, Cistac,
  Rault, Louf, Funtowicz, Davison, Shleifer, von Platen, Ma, Jernite, Plu, Xu,
  Le~Scao, Gugger, Drame, Lhoest, and Rush}]{huggingface}
Thomas Wolf, Lysandre Debut, Victor Sanh, Julien Chaumond, Clement Delangue,
  Anthony Moi, Pierric Cistac, Tim Rault, Remi Louf, Morgan Funtowicz, Joe
  Davison, Sam Shleifer, Patrick von Platen, Clara Ma, Yacine Jernite, Julien
  Plu, Canwen Xu, Teven Le~Scao, Sylvain Gugger, Mariama Drame, Quentin Lhoest,
  and Alexander Rush. 2020.
\newblock Transformers: State-of-the-art natural language processing.
\newblock In \emph{Proceedings of the 2020 Conference on Empirical Methods in
  Natural Language Processing: System Demonstrations}.

\bibitem[{Xie et~al.(2019)Xie, Lai, Doran, and Kadav}]{xie2019visual}
Ning Xie, Farley Lai, Derek Doran, and Asim Kadav. 2019.
\newblock Visual entailment: A novel task for fine-grained image understanding.
\newblock \emph{arXiv preprint arXiv:1901.06706}.

\bibitem[{Yang et~al.(2022)Yang, Li, Zhang, Xiao, Liu, Yuan, and
  Gao}]{yang2022unified}
Jianwei Yang, Chunyuan Li, Pengchuan Zhang, Bin Xiao, Ce~Liu, Lu~Yuan, and
  Jianfeng Gao. 2022.
\newblock Unified contrastive learning in image-text-label space.
\newblock In \emph{Proceedings of the IEEE/CVF Conference on Computer Vision
  and Pattern Recognition}, pages 19163--19173.

\bibitem[{Zhang et~al.(2022)Zhang, Van~Durme, Li, and
  Stengel-Eskin}]{zhang2022visual}
Chenyu Zhang, Benjamin Van~Durme, Zhuowan Li, and Elias Stengel-Eskin. 2022.
\newblock Visual commonsense in pretrained unimodal and multimodal models.
\newblock \emph{arXiv preprint arXiv:2205.01850}.

\bibitem[{Zhao et~al.(2023)Zhao, Zhou, Li, Tang, Wang, Hou, Min, Zhang, Zhang,
  Dong, Du, Yang, Chen, Chen, Jiang, Ren, Li, Tang, Liu, Liu, Nie, and
  Wen}]{LLMSurvey}
Wayne~Xin Zhao, Kun Zhou, Junyi Li, Tianyi Tang, Xiaolei Wang, Yupeng Hou,
  Yingqian Min, Beichen Zhang, Junjie Zhang, Zican Dong, Yifan Du, Chen Yang,
  Yushuo Chen, Zhipeng Chen, Jinhao Jiang, Ruiyang Ren, Yifan Li, Xinyu Tang,
  Zikang Liu, Peiyu Liu, Jian-Yun Nie, and Ji-Rong Wen. 2023.
\newblock \href {http://arxiv.org/abs/2303.18223} {A survey of large language
  models}.
\newblock \emph{arXiv preprint arXiv:2303.18223}.

\bibitem[{Zhu et~al.(2022)Zhu, Yan, Lu, Xu, Wang, Eckstein, and Wang}]{iNLG}
Wanrong Zhu, An~Yan, Yujie Lu, Wenda Xu, Xin~Eric Wang, Miguel Eckstein, and
  William~Yang Wang. 2022.
\newblock Visualize before you write: Imagination-guided open-ended text
  generation.
\newblock \emph{arXiv preprint arXiv:2210.03765}.

\end{thebibliography}
\bibliographystyle{acl_natbib}

\appendix

\newpage

\appendix
\label{appendix}

\section{Ablation Study}
\label{appendix_ablation}

\subsection{Ablation Study on Visual Knowledge Augmentation}

\paratitle{The Effect of the Pre-trained Dataset of VL-PTMs.}  
We notice that the pre-training dataset of VL-PTMs is different from PLMs.
Here, we investigate whether the captions or images from the large-scale image-text pairs contribute more to the performance gain of our approach.
To verify it, we pre-train a new PLM only using the captions data.
Following the setting of ALBEF, we utilize the pre-trained parameters of BERT to initialize this model and only extract the captions from the pre-training data of ALBEF (14.5M sentences in total).
After pre-training on these captions until convergence, we utilize this model to replace CLIP-base in our approach and keep other settings unchanged.
We conduct experiments on commonsense reasoning and NLU tasks to evaluate its effectiveness for augmenting visual knowledge.
As shown in Table~\ref{tableablation1:captions},
we can see that such a variation underperforms ALBEF and our approach, and even leads to performance degradation on the CSQA task. 
It indicates that during pre-training the image data is an important resource for learning visual knowledge in VL-PTMs. 
Only text data (\ie captions) can not provide sufficient visual knowledge that PLMs are hungry for. 
Therefore, after pre-learned on large-scale text-image pairs, CLIP can absorb the useful visual knowledge from the images and inject them into PLMs in our approach.
It further indicates that the improvement of our method is due to the involvement of the visual information about the VH-words.

\begin{table*}[ht]
\small
\centering
\begin{tabular}{cccccc}
\toprule
\textbf{The text encoder of different VL-PTMs~(Params)}  
                                 & \textbf{CSQA-3k} & \textbf{CSQA}         & \textbf{SST-2}  & \textbf{STS-B}  & \textbf{MNLI}   \\
    \midrule
    None                         & 61.59            & 67.90                &  89.23           & 85.46          & 79.06                 \\
    BERT pre-trained on captions~(110M)             & 62.17           &  67.56                &  89.58       & 85.73          & 79.24             \\
    ALBEF~(110M)                 & 63.64            & 68.47               &  90.72           & 87.17          & 80.86            \\
    CLIP-base~(52M)              & \textbf{65.10}   &  \textbf{71.07}       &  \textbf{91.41}  & \textbf{87.73} & \textbf{82.27}              \\
    \bottomrule
\end{tabular}
\caption{Performance comparison of visual representations pre-trained using different pre-training data in our approach. The base model is RoBERTa-base.}
\label{tableablation1:captions}
\end{table*}

\subsection{Ablation Study on Visually-enhanced Fine-tuning}

\paratitle{Different Insertion Positions of Visual Representations.}
In our visually-enhanced fine-tuning framework, we insert the visual representation of the VH-word after its original word embedding.
To verify its effectiveness, we propose three variants of it that do not insert, insert all visual representations of VH-words before and after the input text, respectively.
As shown in Table~\ref{tableablation3:insert position}, we can observe that all these variants would lead to a performance decrease.
It demonstrates that a proper position to insert the visual representation is important for the utilization of augmented visual representations.
By inserting them after the word embeddings of corresponding VH-words, PLMs can effectively aggregate the visual representations to enrich the word representations, leading to better performance on downstream NLP tasks.

\begin{table}[ht]
\centering
\small
\begin{tabular}{cccccccccc}
\toprule
     \multirow{2}{*}{\textbf{Insert Positions}}
    &\multicolumn{4}{c}{\textbf{CSQA-3k}} \\
    \cmidrule(lr){2-5}
    & 5\% & 10\% & 20\% & 100\% \\
    \midrule
    Not insert                & 41.88 & 46.04 & 50.58 & 61.88 \\
    Before input text         & -     & 39.77 & 44.86 & 57.47 \\
    After input text          & - & 40.23 & 45.67 & 58.08 \\
    After the VH-words          & \textbf{42.94} & \textbf{49.27} & \textbf{53.97} & \textbf{65.10} \\
\bottomrule
\end{tabular}
\caption{Performance comparison w.r.t. different insertion positions of visual representations. The base model is RoBERTa-base.}
\label{tableablation3:insert position}
\end{table}

\section{Further Analysis}

\begin{table}[ht]
	\small
	\centering
        \resizebox{0.48\textwidth}{!}{
	\begin{tabular}{cccccc}
	\toprule
	\textbf{}  
								& \textbf{SST-2}   & \textbf{QNLI}  & \textbf{QQP}  & \textbf{STS-B}\\
		\midrule
		  CLIP-base                  & 73.3                   &  74.5       &72.8       &73.8\\
		  Fixed CLIP-base            & 75.1                   &  76.9      &73.7        &75.2\\
		\bottomrule
		
	\end{tabular}}
	\caption{The effect of fixed CLIP's text encoder.}
	\label{tableanalysis:fixedCLIP}
\end{table}

\begin{table}[ht]
\small
\centering     
\begin{tabular}{c c c }
\toprule
   \textbf{ Method}                   & \textbf{Training Time~(s)}     & \textbf{Inference Time~(s)}     \\
    \midrule
    RoBERTa-base   &0.506                   &0.182\\
    +Voken         &0.506                   &0.182\\
    +iACE         &1.138                   &0.512   \\
    +VAWI-SBS      &0.587                   &0.241\\
    +VAWI-VABS      &0.680                   &0.308\\
    +VAWI-LBS      &0.893                   &0.486\\
    \bottomrule
\end{tabular}
\caption{The computation latency during training and inference.} 
\label{tableanalysis:latency}
\end{table}

\paratitle{The Frozen CLIP's Text Encoder.}
In the experiment presented in Table~\ref{tablemain:glue}, we directly fine-tuned CLIP and the results indicate that the performance of VL-PTMs' text encoder is unsatisfactory when directly fine-tuned on NLP tasks.
In our \methodname, we fix the model parameters of CLIP-base's text encoder to preserve the visual knowledge. 
Hence we also conduct experiments on four NLU tasks from GLUE using frozen CLIP. 
Specially, we fix CLIP-base's text encoder and only fine-tuned added 4 transformer layers above it.
As shown in Table~\ref{tableanalysis:fixedCLIP}, we can see that CLIP’s performance under this setting is better than that of directly full-parameter fine-tuning CLIP and also underperforms RoBERTa and BERT.
It indicates that fixing CLIP is more suitable for NLP tasks, and shows the rationality of VAWI settings that always fix the CLIP's text encoder in VAWI to preserve CLIP's knowledge.

\paratitle{The Computation Latency of the Proposed Methods.}
In our \methodname, we fix the model parameters of CLIP-base to preserve the visual knowledge. 
Such a way can also decrease the computation costs during training and inference. 
To verify it, we report the mean training and inference latency per batch on the CSQA-3k dataset of our method and baselines on RTX3090 GPU, where all these methods utilize RoBERTa-base as the backbone. 
As shown in Table~\ref{tableanalysis:latency}, we can see that our proposed VAWI-SBS and VAWI-VABS would not increase the latency too much. 
For VAWI-LBS, as it requires a PLM and a VL-PTM to adaptively select the VH-words, it will relatively increase the latency. 
As shown in Table~\ref{tablemain:glue}, we can see that all the three variants achieve comparable performance in 6 NLU datasets. 
Therefore, it is more efficient and effective to select the SBS and VABS variations in our approach. 
Despite it, we can see that all our variants own less latency than iACE, since our approach does not require a time-consuming image generation process. And as shown in Table~\ref{tablemain:glue}, our approach can also achieve better performance.

\begin{table}[ht]
	\small
	\centering
        \resizebox{0.48\textwidth}{!}{
	\begin{tabular}{cccccc}
	\toprule
	\textbf{Correct VH-words proportions}  
								& \textbf{CSQA-3k}   & \textbf{SST-2}  & \textbf{QQP}\\
		\midrule
		0~\%                    & 61.60                   &  89.57           & 87.63  \\
		20~\%                   & 62.17                   &  89.44           & 87.40  \\
		50~\%                   & 64.22                   &  91.73           & 89.20  \\
		100~\%                  & \textbf{65.10}          &  \textbf{92.93}  & \textbf{89.74}  \\
		\midrule
		None                    & 61.88                   &  89.23           & 86.21  \\
		\bottomrule
		
	\end{tabular}}
	\caption{The effect of the improper visually-hungry words. The base model is RoBERTa-base.}
	\label{robustvhwords}
\end{table}

\paratitle{The Effect of the Improper Visually-hungry Words.}
To analyze how the quality of the VH-words affects the performance of our approach, we further conduct the experiments on CSQA-3K and two NLU tasks SST-2 and QQP from GLUE, to show the effect of insufficient VH-words on our model performance. 
After extracting the VH-words, we remove part of them and only randomly sample 0\%, 20\%, and 50\% VH-words for augmentation. 
As shown in Table~\ref{robustvhwords}, we can see that with the decreasing of the sampling probability, the performance of our approach degrades gradually.
It indicates that not enough VH-words would degrade the performance of our approach.

\begin{figure}[h]
    \centering
    \includegraphics[width=1\columnwidth]{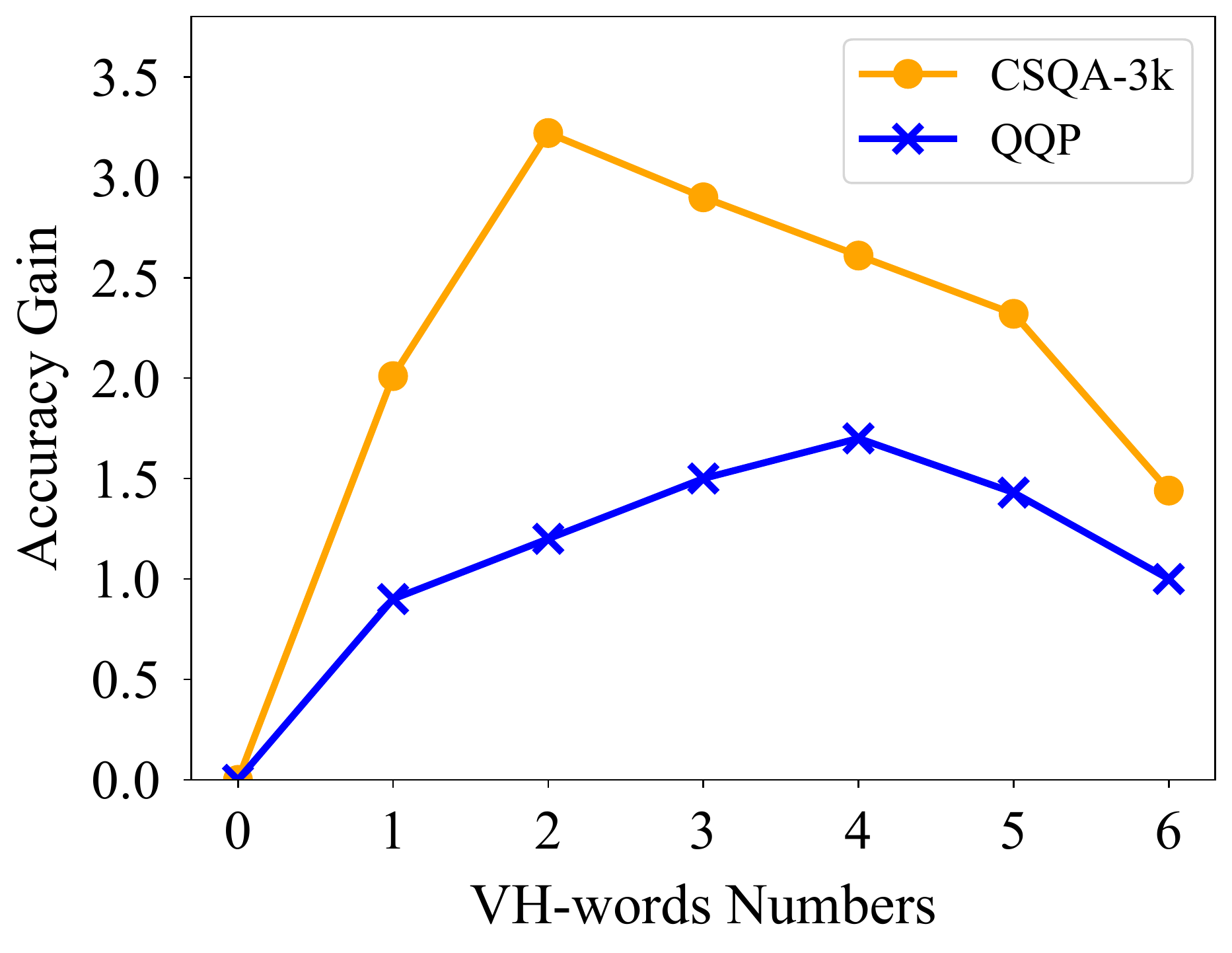}
    \caption{Performance comparison w.r.t. different numbers of VH-words.}
    \label{fig:multi-hop_example}
\end{figure}
\paratitle{The Number of VH-words.}
Our approach has an important hyper-parameter required to tune, such as the number of VH-words.
VH-words can supply visual knowledge that PLMs may be hungry for.
Here, we would like to study whether more VH-words are better to improve performance.
We conduct experiments on the QQP and CSQA-3K datasets using RoBERTa-base as the backbone, and present the results in Figure~\ref{fig:multi-hop_example}.
We can see that with the increase of the number of VH-words, the performance gain of our approach first increases and then decreases.
A possible reason is that too many VH-words may also introduce noisy or redundant information (\eg not very relevant words), which would also influence the fine-tuning performance.
Instead, it is also more efficient to select a few VH-words (\eg two words for CSQA-3k) for deploying our approach in large-scale PLMs.

\paratitle{Case Study of Extracted Visually-hungry Words.}
In this part, we show the VH-words extracted by syntax-, attention- and learning-based strategies in Table~\ref{casestudy:csqa1}, Table~\ref{casestudy:csqa2}, Table~\ref{casestudy:sst-2} and Table~\ref{casestudy:qqp}.
We can see that the three strategies would extract slightly different VH-words. 
The reason is that the three strategies are based on different techniques to identify the VH-words. 
As we can see, the cases show that most of the extracted VH-words by our strategies are generally related to some visual semantics, \eg spider, two eyes.
Although such VH-words can not perfectly cover all the visual semantics, they actually contain most of the important words that the PLMs may be hungry for, \eg red and yellow.
Besides, we can also see that the VH-words extracted by our three strategies may not perfectly align with human judgment. 
In fact, it is also hard for humans to determine proper rules to identify VH-words, \eg people, human, and water. 
In addition, as the learned knowledge of PLM is a black box, it is also difficult for humans to judge the usefulness of our extracted VH-words for PLMs.

\paratitle{The Interpretability of Augmented Embeddings.}
In this part, we show how our augmented embeddings infuse visual knowledge into the PLM.
Concretely, we show the attention distributions of a PLM (\ie RoBERTa-base) in the last few layers before and after infusing visually-augmented representations on CSQA. 
As shown in Table~\ref{AttentionMap}, we can see that the [CLS] tokens pay more attention to the VH-words and their visually-augmented representations, and the VH-words also pay more attention to their visually-augmented representations. 
It shows that the injected visually-augmented representations provide useful knowledge, which guides the PLM to focus on more important tokens and also improves the representations of the VH-words and the [CLS] token.

\begin{table*}[h]
    \small
    \centering
    \begin{tabular}{p{0.97\textwidth}}
        \toprule
        \textbf{Input} \\
        \textit{Input sentence:} Unlike a spider and his many sight seers, people only have what? two eyes.\\
        \midrule
        \textbf{Syntax-based Strategy} \\
        Unlike a \textcolor{greentext}{spider} and his \textcolor{greentext}{many sight seers}, \textcolor{greentext}{people} only have what? \textcolor{greentext}{two eyes} \\
        \midrule
        \textbf{Attention-based Strategy} \\
        Unlike a \textcolor{greentext}{spider} and his many \textcolor{greentext}{sight seers}, \textcolor{greentext}{people} only have what? \textcolor{greentext}{two eyes}. \\
        \midrule
        \textbf{Attention-based Strategy} \\
        Unlike a  \textcolor{greentext}{spider} and his many  \textcolor{greentext}{sight seers},  \textcolor{greentext}{people} only have what?  \textcolor{greentext}{two eyes}.\\
        \bottomrule
        
    \end{tabular}

    \caption{The first instance from the CommonsenseQA dataset. The extracted \textcolor{greentext}{visually-hungry words} are highlighted in green.}
    \label{casestudy:csqa1}
\end{table*}

\begin{table*}[ht]
    \small
    \centering
	\begin{tabular}{p{0.97\textwidth}}
		\toprule
		\textbf{Input} \\
		\textit{Input sentence:} Where on a river can a human hold a cup upright to catch water on a sunny, clear day? waterfall.\\
		\midrule
		\textbf{Syntax-based Strategy} \\
		Where on a \textcolor{greentext}{river} can a \textcolor{greentext}{human} hold a cup upright to catch \textcolor{greentext}{water} on a \textcolor{greentext}{sunny, clear day}? \textcolor{greentext}{waterfall}. \\
		\midrule
		\textbf{Attention-based Strategy} \\
		Where \textcolor{greentext}{on a river} can a \textcolor{greentext}{human} hold a \textcolor{greentext}{cup upright} to catch \textcolor{greentext}{water} on a sunny, clear day? \textcolor{greentext}{waterfall}.\\
		\midrule
		\textbf{Attention-based Strategy} \\
		Where on a \textcolor{greentext}{river} can a human \textcolor{greentext}{hold} a \textcolor{greentext}{cup} upright to \textcolor{greentext}{catch water} on a sunny, clear day? \textcolor{greentext}{waterfall}.\\
		\bottomrule
	\end{tabular}
	\caption{The second instance from the CommonsenseQA dataset. The extracted \textcolor{greentext}{visually-hungry words} are highlighted in green.}
	\label{casestudy:csqa2}
\end{table*}

\begin{table*}[ht]
    \small
    \centering
	\begin{tabular}{p{0.97\textwidth}}
		\toprule
		\textbf{Input} \\
		\textit{Input sentence:} the mesmerizing performances of the leads keep the film grounded and keep the audience riveted.\\
		\midrule
		\textbf{Syntax-based Strategy} \\
		the \textcolor{greentext}{mesmerizing performances} of the leads keep the \textcolor{greentext}{film grounded} and keep the \textcolor{greentext}{audience riveted}.\\
		\midrule
		\textbf{Attention-based Strategy} \\
		the mesmerizing performances of the leads keep the \textcolor{greentext}{film} grounded and keep the \textcolor{greentext}{audience} riveted.\\
		\midrule
		\textbf{Attention-based Strategy} \\
		the mesmerizing performances of the leads keep the film grounded and keep the \textcolor{greentext}{audience riveted}.\\
		\bottomrule
	\end{tabular}
	\caption{The instance from the SST-2 dataset. The extracted \textcolor{greentext}{visually-hungry words} are highlighted in green.}
	\label{casestudy:sst-2}
\end{table*}

\begin{table*}[ht]
    \small
    \centering
	\begin{tabular}{p{0.97\textwidth}}
		\toprule
		\textbf{Input} \\
		\textit{Input sentence:} How do I sell dry Moringa leaves powder in Indian market? Can I use the moringa leaves that are already starting to turn yellow or yellowish?\\
		\midrule
		\textbf{Syntax-based Strategy} \\
		How do I sell \textcolor{greentext}{dry} Moringa \textcolor{greentext}{leaves powder} in Indian \textcolor{greentext}{market}? Can I use the \textcolor{greentext}{moringa leaves} that are already starting to turn \textcolor{greentext}{yellow} or \textcolor{greentext}{yellowish}?\\
		\midrule
		\textbf{Attention-based Strategy} \\
		How do I sell dry Moringa \textcolor{greentext}{leaves powder} in Indian market? Can I use the moringa \textcolor{greentext}{leaves} that are already starting to turn \textcolor{greentext}{yellow} or \textcolor{greentext}{yellowish}?\\
		\midrule
		\textbf{Attention-based Strategy} \\
		How do I sell \textcolor{greentext}{dry} Moringa \textcolor{greentext}{leaves powder} in Indian \textcolor{greentext}{market}? Can I use the moringa \textcolor{greentext}{leaves} that are already starting to \textcolor{greentext}{turn yellow or yellowish}?\\
		\bottomrule
	\end{tabular}
	\caption{The instance from the QQP dataset. The extracted \textcolor{greentext}{visually-hungry words} are highlighted in green.}
	\label{casestudy:qqp}
\end{table*}

\begin{table*}[h]
	\centering
	\small
	\begin{tabular}{c|c}
	\textbf{RoBERTa-Base} & \textbf{\methodname} \\
	\hline
	
	\textbf{Self-Attention, Layer12, Head5} & \textbf{Self-Attention, Layer12, Head5} \\
	\includegraphics[scale=0.4]{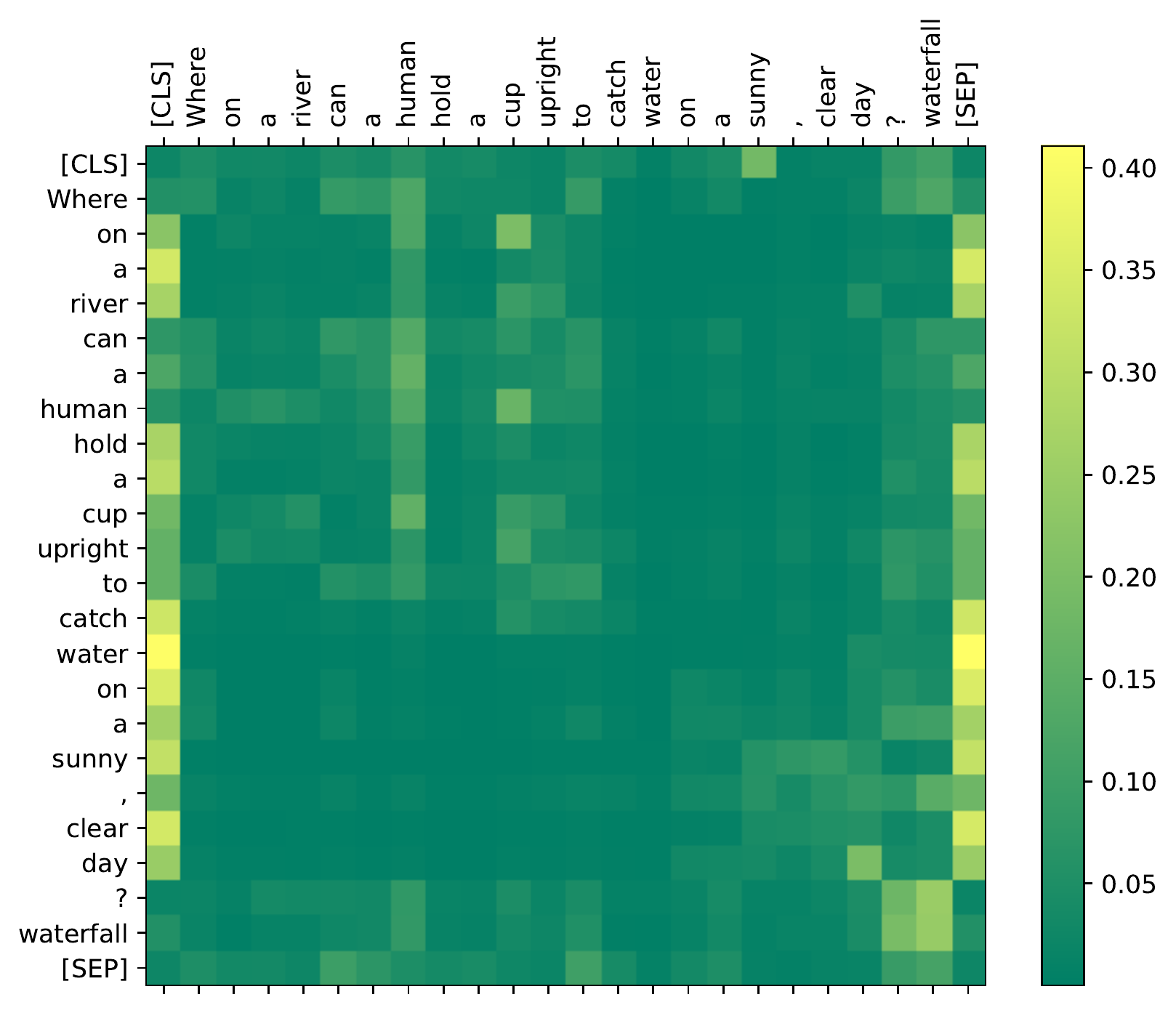} &
	\includegraphics[scale=0.4]{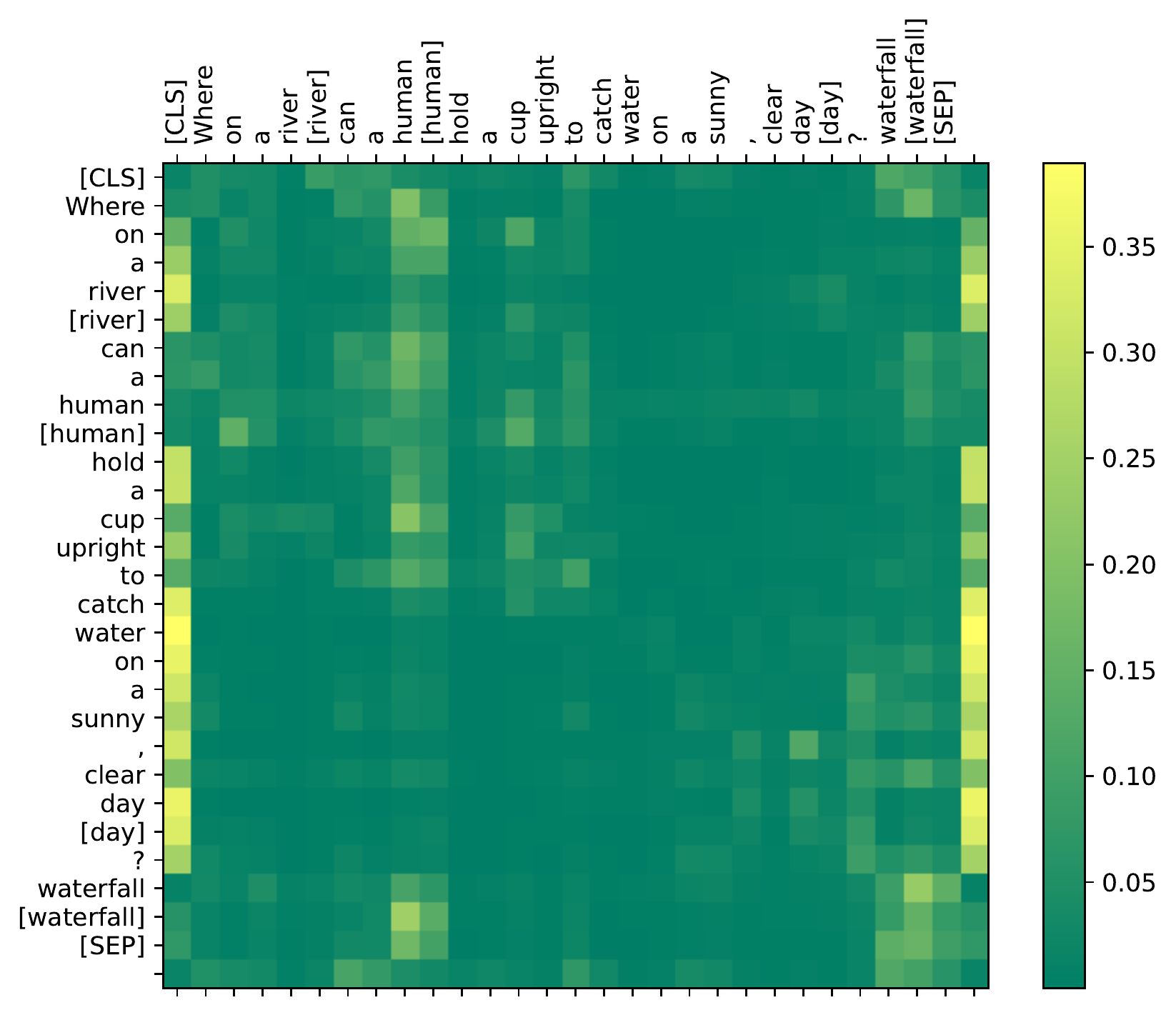} \\
	
	\textbf{Self-Attention, Layer12, Head8} & \textbf{Self-Attention, Layer12, Head8} \\
	\includegraphics[scale=0.4]{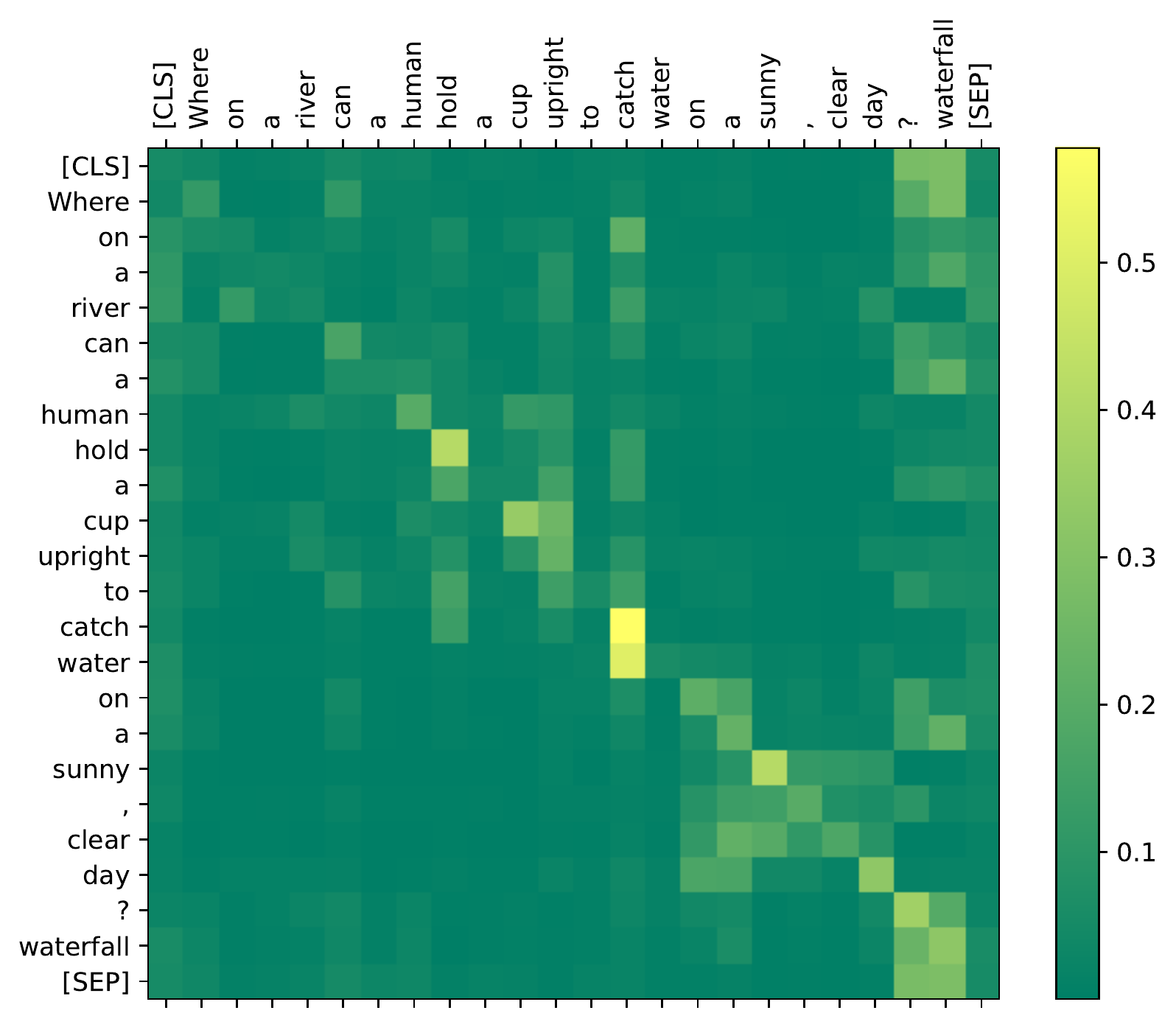} &
	\includegraphics[scale=0.4]{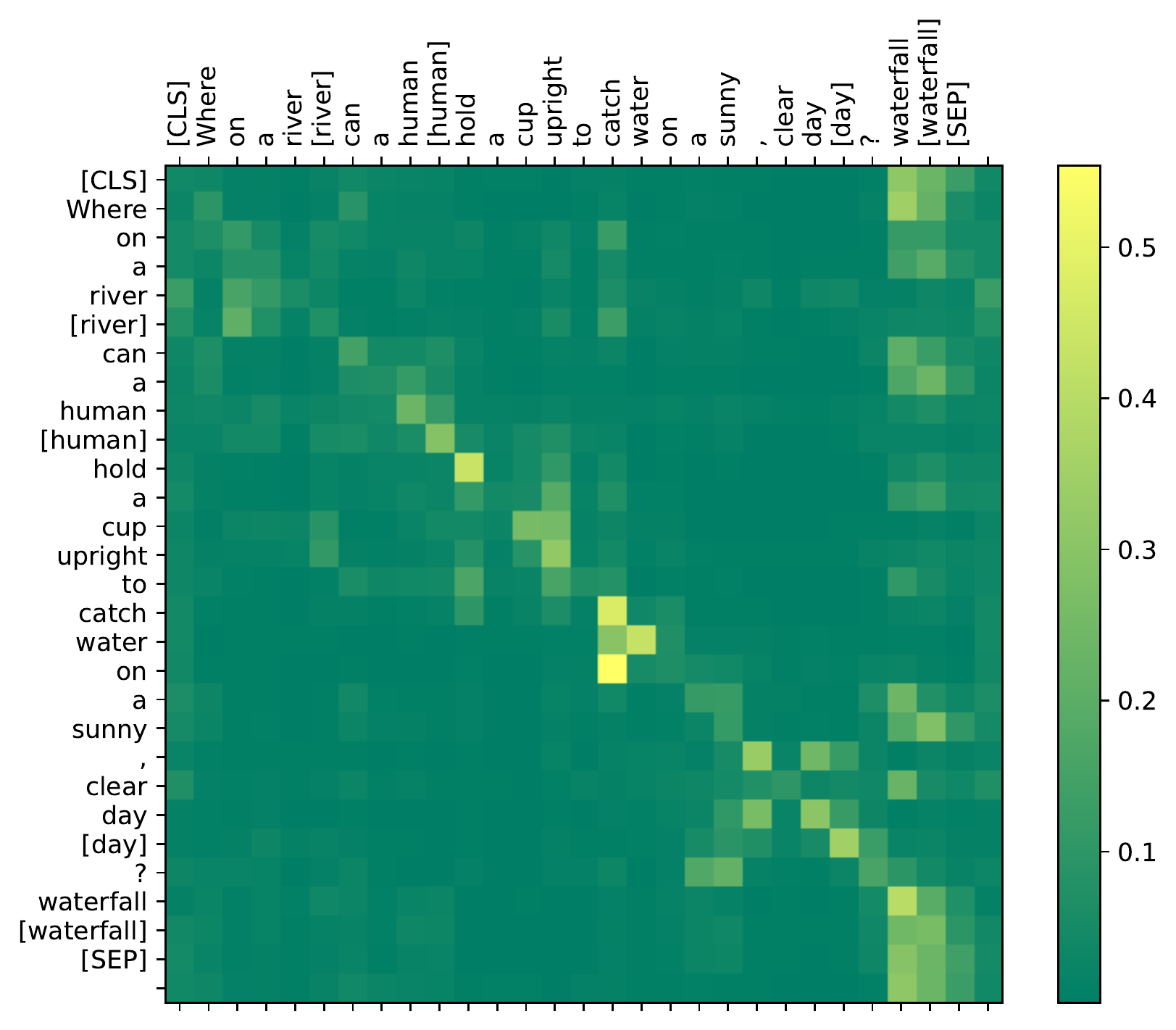} \\
	\end{tabular}
	\caption{The attention maps of the self-attention layers on RoBERTa-base and our approach.}
	\label{AttentionMap}
\end{table*}

\end{document}